\documentclass[10pt, conference, compsocconf]{IEEEtran}
\usepackage[rightcaption]{sidecap}
\usepackage{cite}
\usepackage{amsmath,amsfonts}
\usepackage{array,booktabs}
\usepackage{url}
\usepackage[utf8]{inputenc}
\usepackage{graphicx}
\usepackage{bm}

\providecommand{\B}[1]{\bm{#1}}
\def\slantfrac#1#2{\kern.1em^{#1}\kern-.1em/\kern-.1em_{#2}} 

\begin{document}

\title{On spatial selectivity and prediction across conditions with fMRI}

% \author{
% line 1 (of Affiliation): dept. name of organization\\
% line 2: name of organization, acronyms acceptable\\
% line 3: City, Country\\
% line 4: Email: name@xyz.com
% \and
% line 1 (of Affiliation): dept. name of organization\\
% line 2: name of organization, acronyms acceptable\\
% line 3: City, Country\\
% line 4: Email: name@xyz.com
% }

\author{\IEEEauthorblockN{Yannick Schwartz\IEEEauthorrefmark{1,2},
Ga\"el Varoquaux\IEEEauthorrefmark{1,2},
% Jean-Baptiste Poline\IEEEauthorrefmark{2} 
and Bertrand Thirion\IEEEauthorrefmark{1,2}}
\IEEEauthorblockA{\IEEEauthorrefmark{1}Parietal Team, 
INRIA Saclay-\^{I}le-de-France, Saclay, France\\
Email: yannick.schwartz@inria.fr}
\IEEEauthorblockA{\IEEEauthorrefmark{2}CEA, DSV, 
I\textsuperscript{2}BM, Neurospin b\^{a}t 145, 91191 Gif-Sur-Yvette, France}}

\maketitle              % typeset the title of the contribution

\begin{abstract}
  Researchers in functional neuroimaging mostly use activation
  coordinates to formulate their hypotheses. Instead, we propose to
  use the full statistical images to define regions of interest
  (ROIs). This paper presents two machine learning approaches,
  \emph{transfer learning} and \emph{selection transfer}, that are
  compared upon their ability to identify the common patterns between
  brain activation maps related to two functional tasks. We provide
  some preliminary quantification of these similarities, and show that
  selection transfer makes it possible to set a spatial scale yielding
  ROIs that are more specific to the context of interest than with 
  transfer learning.  In particular, selection transfer outlines well 
  known regions such as the Visual Word Form Area when discriminating 
  between different visual tasks.


\end{abstract}

\section{Introduction}

Functional neuroimaging data are currently routinely used to better 
understand cognitive processes. They rely heavily on previous findings to 
formulate hypotheses and narrow the search space to regions of interest 
(ROIs), most often reported as coordinates of activation peaks 
\cite{yarkoni2011}, or from coordinate databases such as BrainMap 
\cite{laird2005b}. However, understanding the literature is increasingly 
difficult, so that there is a need for more systematic methods, which use the 
images themselves to characterize the functional specificity of brain regions
\cite{sutton2000}. \emph{Transfer learning} is a method that trains a 
classifier to learn a discriminant model on a source task, and then 
generalizes on a target task without further training. It
can yield insights on some brain mechanisms if the tasks 
share specific common effects in some brain regions \cite{knops2009}. 
The goal of this work is to investigate the power of transfer learning
procedures applied to pairs of cognitive contrasts, where the
discrimination ability of the classifier quantifies the information
shared between brain maps, and thus characterizes at which spatial
scale functional contrasts can be jointly classified.
We show that in many cases, transfer learning
gives poor results in terms of spatial selectivity. To address this 
limitation, we introduce \textit{selection transfer}, i.e. classification 
of brain states on the target task following
the canonical procedure \cite{mouraomiranda2005}, but using regions
defined on the source task.

\section{Methods}

\paragraph{Problem setting}

We start from a database holding several studies, each of them containing
different functional contrast images, acquired over multiple subjects. 
We consider two sets of tasks, the \emph{source tasks} and the \emph{target 
tasks}, each composed of pairs of contrast images.
Given $n$ contrasts pairs of $k$ voxel each, we call $\B{X} \in 
\mathbb{R}^{n, k}$ the images of the source tasks, and $\B{y}$ the 
label denoting the functional contrast under study. The target 
images and labels are defined likewise: $\B{X}^{\star} \in \mathbb{R}^{n,k}$ 
and $\B{y}^{\star}$. The source and the target share a similar 
functional \textit{spatial pattern}, and we are interested in finding
the common ROIs, as well as the differences, using a machine learning
approach.
Note that a common pitfall in neuroimaging classification-based data
processing is a successful prediction cannot guarantee that the
information used by the classifier is specific to the cognitive process
of interest.
\paragraph{Regions selection}

Feature selection is an important step of brain activity decoding procedures. Full
brain decoding approaches are efficient but require a careful methodology to
recover the contribution of different brain regions in the classification.
To test the involvement of a particular brain region,
researchers typically use ROIs from an atlas, or derived from the literature.
Another option is to use methods such as the searchlight algorithm, in order to 
evaluate and extract spatially relevant voxels across the whole brain 
\cite{kriegeskorte2006}. We choose to use a one-way ANOVA procedure
\cite{cox2003}, that yields a selection based on the functional activations 
elicited by a task, rather than using purely spatial information. 
We consider different fractions of the brain voxels that are most correlated
to the functional contrast
and perform the learning procedure on these voxels. 
We vary the percentiles of selected voxels with a cubic scale, from
roughly 150 voxels to the full brain. This way we can control the spatial
specificity against the prediction performance, and attempt to find
an optimal set of regions.

\paragraph{Transfer learning}
This consists in learning discriminative models on a \emph{source} functional task 
$(\B{X}, \B{y})$ in order to capture information that should be predictive for
a \emph{target} task $(\B{X^{\star}}, \B{y^{\star}})$. The general
assumption is that if a transfer occurs, the two experiments share at 
least some common cognitive circuity. Here, we train a linear classifier 
on the source task, and we predict the labels of the target without 
any additional training. The features are selected with a one-way ANOVA
on the source task, which makes it possible to compare region-based 
transfer learning with full brain transfer learning.

\paragraph{Selection transfer}
This consists in building 
a predictive model for the target task based on information
extracted from the source task. However, here the transfer occurs on
feature selection: we perform the ANOVA procedure on $(\B{X}, \B{y})$ 
to select the most relevant voxels, then we train a linear classifier on 
$(\B{X^{\star}}, \B{y^{\star}})$, and predict on the same task with the 
voxels selected from the source. Consequently, the 
transfer is not a generalization of a classifier as in transfer learning, but 
rather an evaluation of the significance of features from a task to another.
We use the same linear classifier as the one used for transfer learning.

\section{Experiments and Results}

\subsection{FRMI dataset}

We use data from two fMRI studies for this work. The first one \cite{pinel2007} 
is composed of 322 subjects and was designed to assess the inter-subject 
variability in some language, visual, calculation, and sensorimotor tasks. The
second study is similar to the first one in terms of stimuli, but the data were acquired
on 35 pairs of twin subjects. The two studies were pre-processed and analyzed 
with the standard fMRI analysis software SPM5. The data used for this work are 
a subset of the 90 different statistical images resulting from the
intra-subject analyses. The raw images were acquired on a 3T SIEMENS Trio and a 3T 
Brucker scanner for the first study, and on a 1.5T GE Signa for the second one.
Table \ref{tab:tasks} presents the list of contrasts pairs used for this analysis.

\begin{table*}
\begin{center}

\begin{tabular}{l|cc|cc|l}
  \hline
  Contrasts Names &
  \multicolumn{2}{c|}{ Selected Scale } & 
  \multicolumn{2}{c|}{ Area under p-curve } &
  Description\\
  \hline
  & \emph{trans.} & \emph{sel.} & \emph{trans.} & \emph{sel.} & \\
  \hline
house/scramble $\rightarrow$ face/scramble &68.11 &3.25 &22.73 &4.51& house/scramble = house image versus scrambled image\\
face/scramble $\rightarrow$ house/scramble &0.40 &2.67 &16.22 &2.71& face/scramble = face image versus scrambled image\\
word/scramble $\rightarrow$ face/scramble &23.77 &4.63 &10.36 &2.88& word/scramble = word image versus scrambled image\\
face/scramble $\rightarrow$ word/scramble &1.36 &0.79 &11.15 &2.29& face/scramble = face image versus scrambled image\\
French/sound $\rightarrow$ Korean/sound &0.40 &0.02 &3.57 &4.61& French/sound = French listening versus unstructured sound\\
Korean/sound $\rightarrow$ French/sound &0.27 &0.00 &14.59 &1.21& Korean/sound = Korean listening versus unstructured sound\\
V comp./sent. $\rightarrow$ A comp./sent. &11.01 &0.00 &2.62 &1.76& V comp./sent. = computation versus sentences reading\\
A comp./sent. $\rightarrow$ V comp./sent. &0.01 &6.36 &4.75 &3.10& A comp./sent. = computation versus sentences listening\\
V motor/sent. $\rightarrow$ A motor/sent. &0.10 &0.00 &11.84 &1.85& V motor/sent. = button press action versus sentences reading\\
A motor/sent. $\rightarrow$ V motor/sent. &7.37 &0.00 &4.45 &2.11& A motor/sent. = button press action versus sentences listening\\

\hline
\end{tabular}
\vspace*{-.05in}%
\caption{ Source and target tasks:
\textbf{Selected scales} and \textbf{area under the p-values curve} for 
both transfer learning and selection transfer. trans.= transfer learning; 
sel.= selection transfer; V= visual stimuli; A= auditory stimuli.
\label{tab:tasks}}
\end{center}
\end{table*}

\subsection{Experimental results for transfer learning}

We are interested in \emph{transfer learning}: we learn a discriminative
model on the source task with a univariate feature selection, and predict the 
labels on the target task.

\begin{figure*}[b]

\begin{center}
  \includegraphics[width=1.\linewidth]{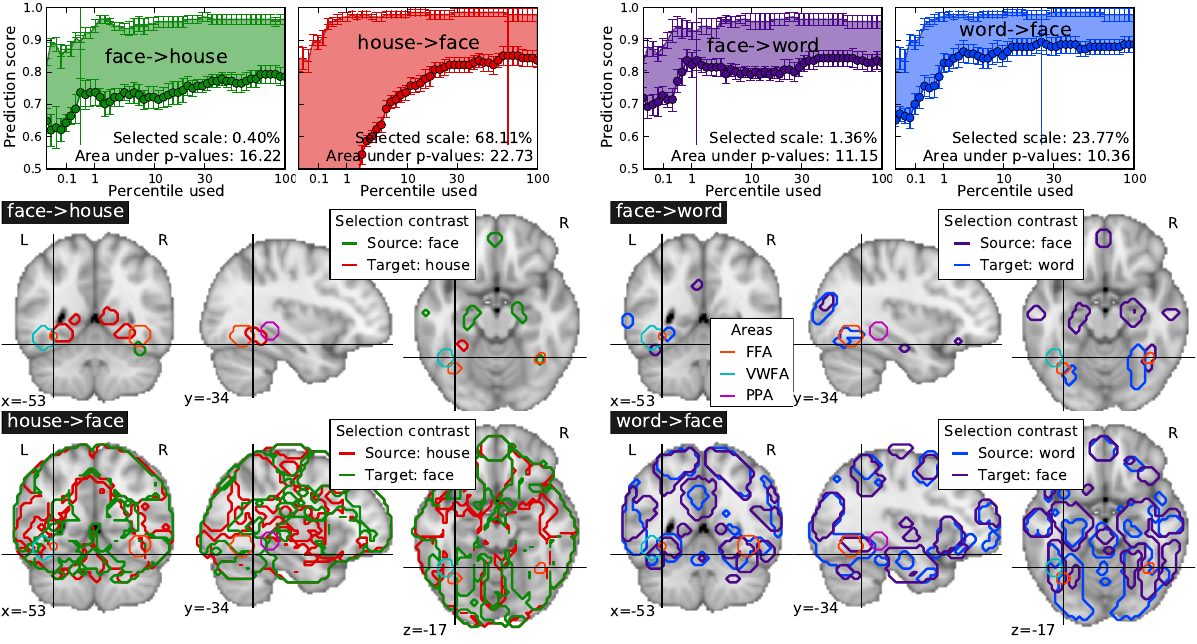}%
\end{center}
\vskip -0.1in
\caption{Example of results using the \textbf{Transfer learning}
approach, in four different transfer settings: we can see that the area
between the inner transfer prediction accuracy curves are large, and that
the prediction rates do not converge. The \textit{optimal scale}, defined
as the minimum of the difference between the curves, often corresponds to
a rather broad, non-specific brain map.
\label{fig:selective}
}
\end{figure*}

The analysis presents two phases: we first train a linear classifier 
on a source task, and then 
re-use the discriminative model on the target task to perform the 
\emph{transfer learning}; this is repeated on 6 different sub-samples of the source 
task to estimate the uncertainty on transfer accuracy. We use two kinds of linear classifiers: a SVC 
(Support Vector Classifier) and a Logistic Regression with $\ell_2$ penalization.
The penalization is set by nested 
6-fold cross-validation for each classifier. We find that the two methods yield very close results, and thus 
report only results using the SVC classifier. We also train and then test the 
classifier on the target task and call this procedure \emph{inline learning}. 
In Figure \ref{fig:selective}, we show the performance $\tau^{t}_{p}$ of
transfer learning, relative to inline learning $\tau^{i}_{p}$, varying 
the percentile $p$ of features selected in a cubic scale. In general, 
for any given $p$, $\tau^{i}_{p}$ can remain  
significantly higher than $\tau^{t}_{p}$. For this reason, we use a 
heuristic to select the scale parameter (see also Figure \ref{fig:selective}):
the scale that yields the minimal $\tau^{i}_{p}-\tau^{t}_{p}$ difference. We consider that at this 
scale, the maps associated with the two tasks share a maximal amount of common information. 

However, the voxels selected with this method are either too few to give an accurate
prediction, or too many to yield identifiable regions. The transfers
do not behave the same way on both directions: in general, one direction is
more sensitive but less specific, and the other direction shows the opposite behaviour.
This comes from tasks-related foci being more spatially focused for some contrasts. Because
of this lack of specificity, we do not find contained regions that overlap
with the Fusiform Face Area (FFA) \cite{kanwisher1997fusiform}, the Parahippocampal 
Place Area (PPA) \cite{epstein1998} or the Visual Word Form Area (VWFA)
\cite{cohen2004}, regions respectively involved in face recognition, object 
visual processing, and reading.

\subsection{Experimental results for selection transfer}

We are interested in \emph{selection transfer}: we do not perform transfer
learning, instead, we use the univariate feature selection performed on the
source task, to learn a discriminative model and predict the labels in the
target task.

We use the same machine learning tools as the transfer learning: we train and
test a linear classifier with a 6-fold cross validation test on the 
target task. For this method the SVC and the Logistic Regression 
with $\ell_2$ penalization also give very close results. As with 
\emph{transfer learning}, we also perform an inline learning on the 
target task, with features selected on the same images.

\begin{figure*}[t]

\begin{center}
  \includegraphics[width=1.\linewidth]{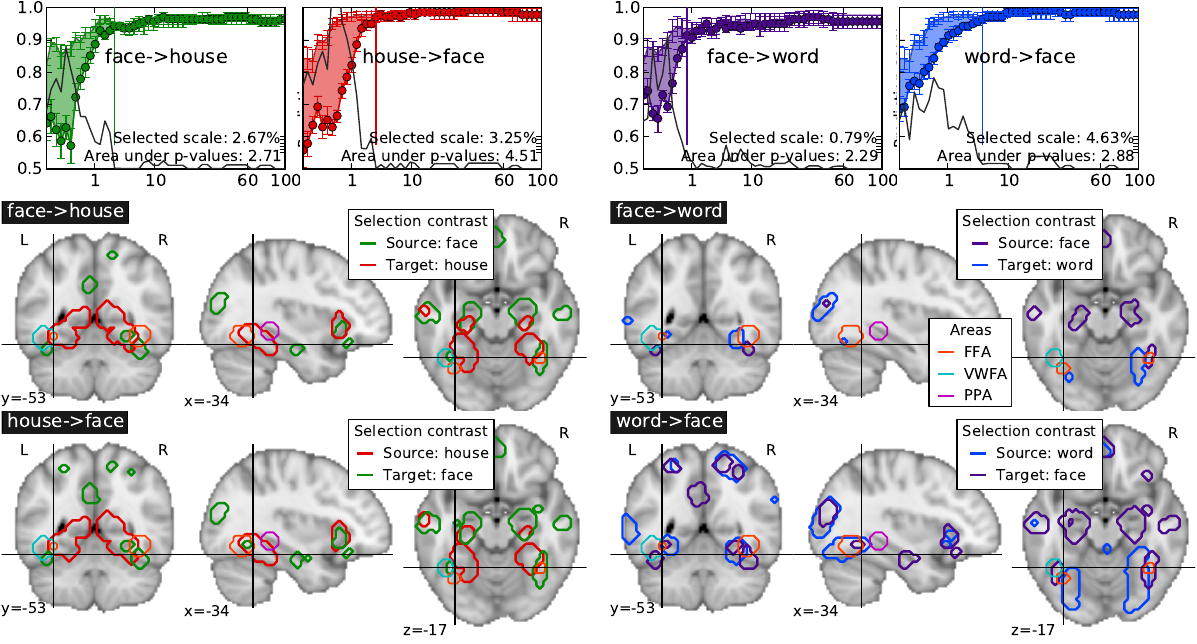}%
\end{center}
\vskip -0.1in
\caption{ Example of result using the \textbf{Selection transfer}
  approach: The two prediction curves do converge, so that the
  difference becomes non-significant as soon as a relatively small
  fraction of the voxels are included: the spatial scale is defined
  here as the point where the curves can no longer be distinguished. It
  corresponds to more symmetric and meaningful brain maps than those obtained with
  transfer learning.
\label{fig:specific}
}
\end{figure*}

On Figure \ref{fig:specific}, we show the performance $\tau^{s}_{p}$ of 
selection transfer against inline learning $\tau^{i}_{p}$, and how the 
performance varies with the percentile $p$ of the brain recruited for the 
learning process. In comparison to transfer
learning, two things happen: \emph{i)} the selection transfer is more
symmetric, \emph{ii)}
$\tau^{i}_{p}$ is not significantly higher than $\tau^{s}_{p}$ for every $p$.
We can therefore use a t-test to define the \emph{selected scale} 
(Figure \ref{fig:specific}) as the first one with non significant 
difference between the curves. This enables us to control 
the amount of information to include in the prediction problem, and have both
a good performance and an improved specificity of the regions selected
for the two tasks. In practical terms, the \emph{selected scale} makes it possible to 
identify the smallest fraction of the brain that yields overlapping regions 
in the two tasks, and consequently an accurate prediction.
Although the selected regions have no guarantee of optimality, they 
are specific enough to overlap with the FFA, the PPA and the VWFA. 
We can also use the area under the p-values curve from the t-test as a measure
of similarity between the tasks. While, it is not possible to interpret this 
measure absolutely, we can use it to compare one task versus others.
For the example on Figure \ref{fig:specific}, we can see that the area
between \emph{face} and \emph{word} is smaller than between \emph{face} and
\emph{house}. This indicates that the face task is closer to the word task
than the house task, which is consistent with previous findings \cite{dehaene2010}.

\paragraph{Limitations}

\emph{Selection transfer} captures voxels that generalize well in terms of 
prediction from one task to another. However, a classifier may require very 
few voxels to perform well, in which case this method misses some
regions involved in the cognitive process of interest. This effect is
represented by the values in Table \ref{tab:tasks}, where \emph{selection
transfer} requires only a small $p$ fraction of the brain to obtain a $\tau^{s}_{p}$, 
which is not significantly lower than $\tau^{i}_{p}$ (e.g., V comp./sent 
$\rightarrow$ A motor/sent.). In order to retrieve optimal regions when 
this is the case, a standard analysis, based either on contrast addition 
or conjunction \cite{nichols2005valid}, would be sensitive enough to detect 
the common active regions for both tasks.

\section{Conclusion}

In this contribution, we investigate the ability of \emph{transfer learning}
and \emph{selection transfer} to characterize the spatial scale at which functional
contrasts can be jointly classified. The objective is to find a systematic
procedure to extract ROIs that define common information between two functional
tasks, instead of relying on activation coordinates from the literature.
We show that transfer learning does not provide control on the regions size it 
uses to classify the tasks. Instead we use a selection transfer procedure that seems
to better characterize which fraction of the brain yields discriminant information. 
Our results suggest that transfer learning requires to be used in a carefully
designed study, as it is difficult to control the spatial selectivity of this method.
Another interesting result is that selection transfer is not symmetric (i.e., source 
and target tasks are not inversible), as opposed to contrast conjunction. In the 
future, we would like use such methods in meta-analysis, in order to leverage 
large databases of functional images.

\section{Acknowledgements}

This work was supported by the ANR grants BrainPedia ANR-10-JCJC 1408-01
and IRMGroup ANR-10-BLAN-0126-02.

\bibliography{biblio}

\end{document}